\documentclass[letterpaper, 10pt, conference]{ieeeconf}  

\IEEEoverridecommandlockouts                              

\usepackage{graphics} 
\usepackage{amsmath} 
\usepackage{amssymb}  

\usepackage{algorithm}
\usepackage{algpseudocode}
\usepackage{varwidth}

\usepackage{url}

\usepackage[font=footnotesize]{caption}
\usepackage{subcaption}
\usepackage{tikz}
\usepackage{gensymb} 
\usepackage{bm} 
\usepackage{graphbox}  

\usepackage[hidelinks]{hyperref}  

\algrenewcommand\algorithmicindent{0.6em}

\def\equationautorefname~#1\null{Equation (#1)\null}  

\usepackage{color}
\usepackage{amsfonts}
\usepackage{ifthen}
\usepackage{xifthen}

\newcommand{\etal}{{et~al.\ }}

\newcommand{\expectation}{{\mathbb{E}}}

\newcommand{\entropy}{{\mathcal{H}}}
\newcommand{\gauss}{{\mathcal{N}}}

\newcommand{\Real}{\mathbb{R}}
\newcommand{\trsp}{\mathsf{T}}  

\newcommand{\dimension}[1][]{%
	\ifthenelse{\isempty{#1}}%
	{{\text{D}}}
	{{\text{D}_#1}}
}

\newcommand{\mdp}{{\mathcal{M}}}
\newcommand{\mmdp}{{\mathcal{M}}}  
\newcommand{\tmdp}{{\mathcal{T}}}  
\newcommand{\tmdpSet}{{\mathcal{T}}}  

\newcommand{\state}{{\mathbf{s}}}
\newcommand{\sspace}{\mathcal{S}}
\newcommand{\st}{{\mathbf{s}_t}}
\newcommand{\stp}{{\mathbf{s}_{t+1}}}

\newcommand{\sdim}{{\dimension[\sspace]}}

\newcommand{\action}{{\mathbf{a}}}
\newcommand{\aspace}{\mathcal{A}}
\newcommand{\at}{{\mathbf{a}_t}}

\newcommand{\adim}{{\dimension[\aspace]}}
\newcommand{\reward}{r}

\newcommand{\rt}{\reward_t}

\newcommand{\rvect}{\mathbf{\reward}}
\newcommand{\rvectt}{\mathbf{\reward}_t}
\newcommand{\return}{{G}}

\newcommand{\discount}{{\gamma}}
\newcommand{\pdyn}{{p_s}}

\newcommand{\policy}{{\pi}}

\newcommand{\pparams}{{\boldsymbol\theta}}   
\newcommand{\ppolicy}{{\policy_\pparams}}   

\newcommand{\qparams}{{\boldsymbol\phi}}   
\newcommand{\pqval}{{Q_{\qparams}}}
\newcommand{\traj}{{\tau}}

\newcommand{\trajinfinite}{{\traj = \{\state_0, \action_0, \state_1, \action_1, \dots\}}}
\newcommand{\rbuffer}{{\mathcal{D}}}  

\newcommand{\demopolicy}[1][]{%
	\ifthenelse{\isempty{#1}}%
	{{\policy^{\text{E}}}}
	{{\policy_{#1}^{\text{E}}}}
}
\newcommand{\demoppolicy}[1][]{%
	\demopolicy[\pparams]
}
\newcommand{\demotraj}[1][]{%
	\ifthenelse{\isempty{#1}}%
	{{\traj^{\text{E}}}}
	{{\traj_{#1}^{\text{E}}}}
}
\newcommand{\demopolicyJ}[1][]{%
	\ifthenelse{\isempty{#1}}%
	{{\policy^{\text{E}^{(j)}}}}
	{{\policy_{#1}^{\text{E}^{(j)}}}}
}
\newcommand{\demoppolicyJ}[1][]{%
	\demopolicyJ[\pparams]
}
\newcommand{\demotrajJ}[1][]{%
	\ifthenelse{\isempty{#1}}%
	{{\traj^{\text{E}^{(j)}}}}
	{{\traj_{#1}^{\text{E}^{(j)}}}}
}

\newcommand{\task}{\mathcal{T}}
\newcommand{\dataset}{\mathcal{D}}

\newcommand{\higherpol}{{\policy^{\mdp}}}

\newcommand{\higherpolfcn}{{\mathtt{f}}}

\newcommand{\compopol}{{\policy^{[k]}}}

\newcommand{\compopolSet}{{\Pi}}

\newcommand{\activationvect}{\mathbf{w}^{[k]}}
\newcommand{\activationfcn}{{\mathtt{h}}}
\newcommand{\activationSet}{{\mathcal{W}}}

\title{\LARGE \bf
  Hierarchical Reinforcement Learning for Concurrent Discovery of Compound and Composable Policies
}

\author{Domingo Esteban$^{1,2}$, Leonel Rozo$^{3}$ and Darwin G. Caldwell$^{1}$
\thanks{$^{1}$Department of Advanced Robotics, Istituto Italiano di Tecnologia, Via Morego 30, 16163 Genova, Italy
  {\tt\small domingo.esteban@iit.it} and {\tt\small darwin.caldwell@iit.it}}
\thanks{$^{2}$DIBRIS, Universit\`a di Genova, Via Opera Pia 13, 16145, Italy}
\thanks{$^{3}$Bosch Center for Artificial Intelligence, Robert-Bosch-Campus 1, 71272 Renningen, Germany. {\tt\small leonel.rozo@de.bosch.com}}
}

\begin{document}

\maketitle
\thispagestyle{empty}
\pagestyle{empty}

\begin{abstract}
A common strategy to deal with the expensive reinforcement learning (RL) of complex tasks is to decompose them into a collection of subtasks that are usually simpler to learn as well as reusable for new problems.
However, when a robot learns the policies for these subtasks, common approaches treat every policy learning process separately.
Therefore, all these individual (composable) policies need to be learned before tackling the learning process of the complex task through policies composition.
Moreover, such composition of individual policies is usually performed sequentially, which is not suitable for tasks that require to perform the subtasks concurrently.
In this paper, we propose to combine a set of composable Gaussian policies corresponding to these subtasks using a set of activation vectors, resulting in a complex Gaussian policy that is a function of the means and covariances matrices of the composable policies.
Moreover, we propose an algorithm for learning both compound and composable policies within the same learning process by exploiting the off-policy data generated from the compound policy.
The algorithm is built on a maximum entropy RL approach to favor exploration during the learning process.
The results of the experiments show that the experience collected with the compound policy permits not only to solve the complex task but also to obtain useful composable policies that successfully perform in their corresponding subtasks.
\end{abstract}

\section{Introduction}\label{sec:Introduction}
Reinforcement learning (RL) is a general framework that allows an agent to autonomously discover optimal behaviors by interacting with its environment.
However, when applied to robotics scenarios some specific challenges arise, such as high-dimensional continuous state-action spaces, real-time requirements, delays and noise in sensing and execution, and expensive (real-world) samples~\cite{peters16:robotlearning}. Recently, several authors have overcome some of these challenges by using deep neural networks (NN) as parameterized policies for generating rich behaviors in end-to-end frameworks~\cite{lillicrap16:ddpg}\cite{levine16:end-to-end_deep_visuomotor_pols}. Nevertheless, learning the high number of parameters of deep NNs in complex tasks involves a large number of interactions with the environment that compromises the real-world sample challenge.

\begin{figure}[!htb]
  \vspace*{2ex}
  \centering
  \includegraphics[width=.98\linewidth]{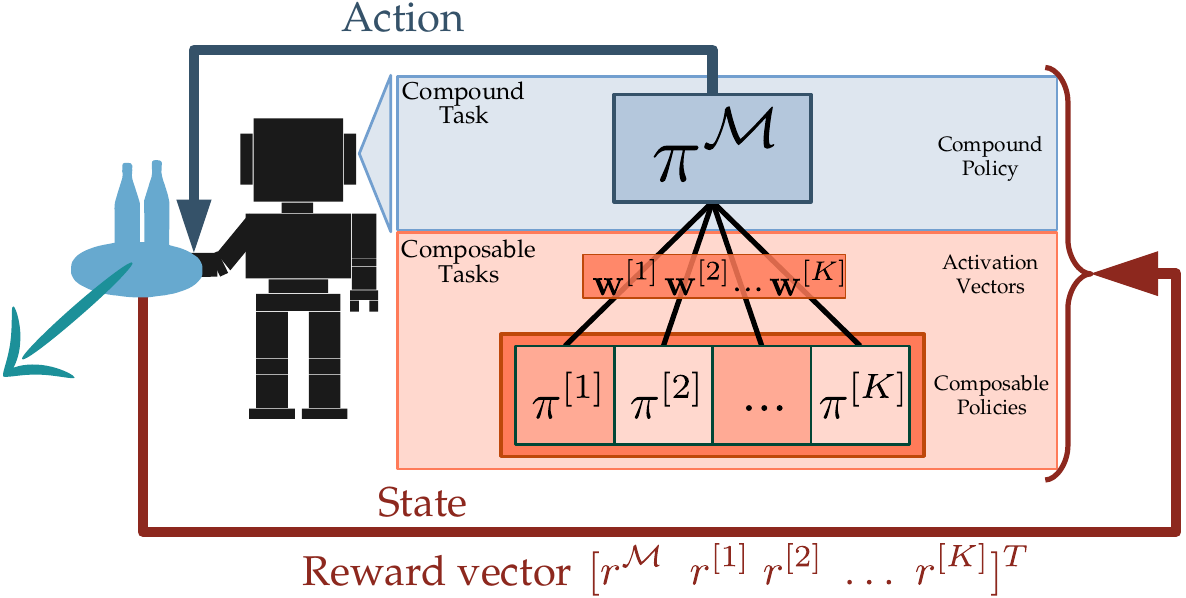}
  \caption{
    A possibly complex robotic task (\textit{compound} task) can be decomposed into a collection of simpler and reusable tasks (\textit{composable} tasks). During the RL process, the robot interacts with the environment and receives a vector with the corresponding rewards. At each iteration, the robot action is sampled from the compound policy $\higherpol$ obtained from the combination of the corresponding composable policies $\{\compopol\}_{1}^{K}$ and a set of state-dependent activation vectors $\{\activationvect\}_{1}^{K}$. Both composable policies and activation vectors are learned simultaneously using the same interaction data.
    }
  \label{Fig:hiu_diagram}
  \vspace*{-2ex}
\end{figure}

Several algorithms have been proposed to improve sample efficiency of model-free deep RL by making a better use of the sample information (data-efficiency), obtaining more information from data (sample choice) and improving several times the policy with the same samples (sample reuse)~\cite{sigaud18:ps_continuous_action_domains}.
However, learning a complex robotic task may still be slow or even infeasible when the robot learns from scratch.
An appealing approach to deal with this problem is to decompose the task into subtasks that are both simpler to learn and reusable for new problems.

Many tasks in robotics may intuitively be divided into individual tasks, for example, the task of moving an object to a specific location may be decomposed into reaching for the object, grasping it, moving it to the target, and releasing it.
Therefore, when a robot is provided with a collection of policies defined in these \textit{composable} tasks, a new RL problem can be stated as learning how to mix these composable policies such that the performance criterion of the complex task is optimized.
Note that shifting the complexity of this task learning problem to layers of simpler functionality is mainly studied in the field of hierarchical RL.
However, the assumptions of common hierarchical RL algorithms limit their application in several robotic tasks.

Firstly, several hierarchical RL methods decompose the complex RL problem temporally, meaning that during a certain period of time the behavior of the robot is delegated to a specific subtask policy~\cite{daniel16:hreps-journal}.
This temporal decomposition is suitable for robotic tasks that can be divided sequentially, however, many others require the robot to perform individual tasks concurrently.
For example, a manipulator carrying an object and avoiding an obstacle at the same time.
Secondly, common learning methods optimize the individual policies and the compound policy through independent single-task RL processes
~\cite{muelling10:momp}%
\cite{osa18:hrl_multiple_grasping_strategies_human_instructions}%
\cite{haarnoja18:composable_drl}%
.
Therefore, the robot has to continuously interact with the environment to learn, possibly from scratch, first a collection of composable policies, and only after that their composition, compromising sample efficiency.

Under this scenario, we propose a two-level hierarchical RL approach where a set of Gaussian policies, constituting the low level of the hierarchy, are composed at the high level by means of state-dependent activation vectors defined for each policy.
These activation vectors allow to consider concurrently actions sampled from all the low-level policies and preferences among specific components.
Furthermore, we propose in \autoref{sec:composition_max_ent_policies} two alternatives for obtaining a compound Gaussian policy as a function of the parameters of the low-level policies and their corresponding activation vectors.
An algorithm for learning both high- and low-level policies within the same learning process is proposed and detailed in \autoref{sec:simultanous_learning_and_compo}.

As an illustration, \autoref{Fig:hiu_diagram} shows how the proposed algorithm executes actions sampled from the high-level policy, and then exploits this experience for learning simultaneously both low-level policies and activation vectors in an off-policy manner.
The results of the experiments detailed in \autoref{sec:Experiments} show that the proposed algorithm obtains a high-level policy that solves compound tasks while learning useful low-level policies that can be potentially reused for new tasks.
Supplementary videos and code of the proposed approach are available at: {\footnotesize \url{https://sites.google.com/view/hrl-concurrent-discovery}}

\section{Related Work}\label{RelatedWork}
Complex problems in RL usually involve either tasks that can be hierarchically organized into subtasks, scenarios requiring concurrent execution of several tasks, and tasks with large or continuous state-action spaces~\cite{sprague03:multigoalRLwithModularSARSA}.
Hierarchical RL approaches split a complex task into a set of simpler elementary tasks~\cite{barto03:survey_hrl}.
Some of these methods have been successfully applied in robotics by exploiting temporal abstraction, where the decision to invoke a particular task is not required at each time step but over different time periods~\cite{daniel16:hreps-journal}\cite{kober11:learning_elementary_movs_jointly_higher_level_task}.
As a consequence, these methods assume that a high-level policy, which selects the subtask, and low-level policies, which select the action, are executed in different time scales.
In contrast, the approach proposed in this paper considers that the decisions at both levels of the hierarchy are executed at each time step.

The temporal abstraction assumption in most hierarchical RL methods also involves that during a certain period of time the robot only performs a particular task.
RL problems requiring policies that solve several tasks at the same time are commonly stated as multiobjective or modular RL problems~\cite{liu15:multiobjectiveRL}\cite{simpkins18:composable_modular_RL}.
The policies of all these subtasks may be combined using weights describing the predictability of the environmental dynamics~\cite{doya02:multiple_model-based_rl}, or the values obtained from the desirability function in a linearly-solvable control context~\cite{uchibe14:combining_learned_controllers}.
Another alternative is to combine action-value functions of composable tasks, and then extract a policy from this combined function~\cite{haarnoja18:composable_drl}.
The latter paper is similar to ours since the composable policies are also optimizing an entropy-augmented RL objective, however, their combination is carried out at the level of action-value functions unlike our policy-based approach.
Moreover, their composable policies have been previously learned in independent processes, in comparison to the algorithm proposed here where both compound and composable policies are improved in the same RL process.

Exploiting the experience collected using a particular policy (so-called behavior policy) to concurrently improve several policies is a promising strategy that has been previously explored in literature.
The Horde architecture shows how independent RL agents with same state-action spaces can be formulated for solving different problems~\cite{sutton11:horde}.
This approach proposed by Sutton \etal is relevant because exploits its off-policy formulation for improving all the policies in parallel.

Several works extended the Horde formulation and used deep NN for dealing with high dimensional or continuous state-action spaces, and also to provide a modularity that can be exploited for modeling and training independent RL problems~\cite{cabi17:iu_agent}\cite{schaul15:uvfa}\cite{jaderberg16:rl_unsup_aux_tasks}\cite{yang18:hierarchical_drl_continuous_action}.
The Intentional-Unintentional agent~\cite{cabi17:iu_agent}, for example, shows how deep NN policies and value functions can be learned even with an arbitrarily selected behavior policy.
However, this policy can also be chosen at each time step by following a hierarchical objective, and the selection process can be improved as a function of the performance in the complex task~\cite{yang18:hierarchical_drl_continuous_action}.
Note that the policies in~\cite{cabi17:iu_agent}\cite{yang18:hierarchical_drl_continuous_action} are deterministic, and then the exploration should be carried out by adding a noise generated from an Ornstein-Uhlenbeck process.
In contrast, all the policies in this paper are stochastic and the exploration is generated directly from the behavior policy, that is the stochastic high-level policy.

Most of the notation and the approach proposed in this paper are inspired by the Scheduled Auxiliary Control (SAC-X) method~\cite{riedmiller18:sac-x}.
SAC-X solves complex tasks based on a collection of simpler individual tasks, and learns from scratch, both high- and low-level policies simultaneously.
However, this method considers temporal abstraction in the hierarchy, and therefore the high-level policy is a scheduler that occasionally selects one low-level policy.
Therefore, the policies at the low level of the hierarchy can only be executed sequentially and run at a different time-scale than the policy defined in the high level.
This methodology differs from our approach as we consider a framework that is able to execute different policies concurrently.

\section{Preliminaries}\label{sec:Preliminaries}

The sequential decision making process of a robot could be modeled by a Markov decision process (MDP) $\mdp$, defined by the tuple $(\sspace, \aspace, \pdyn, \reward)$, where $\sspace \subset \Real^\sdim$ and $\aspace \subset \Real^\adim$ are continuous state and action spaces of dimensionality $\sdim$ and $\adim$, respectively.
At each time step $t$, the robot selects an action $\at \in \aspace$ according to a policy $\policy$ which is a function of the current state of the environment ${\st \in \sspace}$.
After this interaction, the state of the environment changes to $\stp \in \sspace$ with a probability density $\pdyn = p(\stp|\st, \at)$, and the robot receives a reward according to the function ${r: \sspace \times \aspace \rightarrow \Real}$.
The robot's goal is to maximize the expected infinite-horizon discounted accumulated reward $\return(\traj)$ of the trajectory $\trajinfinite$, induced by its policy $\policy$.
That is to say ${J(\policy) = \expectation_{\traj}\;[\return(\traj)] = \expectation_{\traj}\;[\sum_{t=0}^{\infty} \discount^{t}\;  r(\st, \at)]}$.


\subsection{Maximum Entropy Reinforcement Learning}
\label{sec:max_ent_rl}
The exploration required for a robot to generate behaviors that produce high return can be directly obtained by the stochastic policy $\policy(\at|\st)$, where the randomness of the actions $\at$ given the state $\st$ can be quantified by the entropy of the policy.
\emph{Maximum entropy reinforcement learning} or entropy regularized RL \cite{ziebart10:maxent-thesis}\cite{neu17:unified_view_entro_regu_mdp} is a formulation that augments the previous RL objective by including the entropy of the robot's policy 
${J(\policy) = \expectation_{\traj} \left[ \sum_{t=0}^{\infty} \discount^t \Big( r(\st, \at) + \alpha \entropy(\policy(\cdot | \st)) \Big) \right]}$, where $\entropy(\policy(\cdot | \st))$ denotes the entropy of an action $\at$ with distribution $\policy(\at | \st)$ and is computed as ${\entropy(\policy(\cdot | \st)) = \expectation_{\at \sim \policy} [- \log \policy(\cdot | \st)]}$.
The parameter $\alpha$ controls the stochasticity of the optimal policy.
Note that the conventional RL objective is recovered in the limit $\alpha \rightarrow 0$. 

\subsection{Soft Actor-Critic algorithm}
\label{sec:sac}

\emph{Soft actor-critic} (SAC) is an off-policy actor-critic deep RL algorithm that optimizes stochastic policies defined in the maximum entropy framework \cite{haarnoja18:sac}.
The algorithm is built on a policy iteration formulation that alternates between policy evaluation and policy improvement steps.
In the former, a parameterized soft Q-function $\pqval$ is updated to match the value of the parameterized policy $\ppolicy$ according to the maximum entropy objective, while in the latter the policy $\ppolicy$ is updated towards the exponential of the updated $\pqval$.
The update rules for the NN parameters $\qparams$ and $\pparams$ are given in \autoref{sec:simultanous_learning_and_compo}, and more details of the SAC algorithm can be found in \cite{haarnoja18:sac}\cite{levine18:rl_and_control_as_prob_inference}.

\section{Composition of modular Gaussian policies}%
\label{sec:composition_max_ent_policies}
Acquiring autonomously a robotic skill for a specific task can be achieved by directly optimizing the maximum entropy RL objective with the experience collected from the task execution.
Nevertheless, when a new task defined in the same state and action spaces has to be learned, the robot should interact again with the environment to obtain useful data for improving the policy for this new task.
Sample efficiency is a key concern in robotics, therefore in this section we propose a two-level hierarchical model that seeks to construct a set of simpler and reusable policies in the low level of the hierarchy, and a high-level policy that combines them for solving a more complex task.
In addition, in section \ref{sec:simultanous_learning_and_compo}, we will introduce an algorithm for exploiting better the interaction data and learning both low- and high-level policies all together.

\subsection{Hierarchical model for composing modular policies}\label{sec:hierarchical_model}
First, let us assume that several complex tasks can be decomposed into a set of $K$ \emph{composable tasks} ${\tmdpSet = \{\tmdp^{[k]}\}_1^K}$.
All of them have the same state space, action space and transition dynamics, however, each one is characterized by a specific reward function  $\reward^{[k]}(\st, \at)$.
Thus, the corresponding MDP for each composable task $\tmdp^{[k]}$ is ${(\sspace, \aspace,  \pdyn, \reward^{[k]})}$.
Second, let us assume that stochastic policies defined in these MDPs, called \emph{composable policies}, are conditional Gaussian distributions ${\compopol(\action|\state) = \gauss(\mathbf{m}^{[k]}, \mathbf{C}^{[k]})}$, with mean vector $\mathbf{m}^{[k]} \in \aspace$ and diagonal covariance matrix ${\mathbf{C}^{[k]} = \text{diag}[\sigma^2_1,\sigma^2_2,\dots,\sigma^2_\adim]}$. 

Let us also define a (possibly more complex) \emph{compound task} $\mmdp$, described by the combination of the tasks in $\tmdpSet$.
As with the composable tasks, $\mmdp$ also shares the same state space, action space and transition dynamics, but it is characterized by the reward $\reward^{\mdp} (\st, \at)$.
Finally, a stochastic policy defined in the resulting MDP ${(\sspace, \aspace,  \pdyn, \reward^{\mdp})}$ is named \emph{compound policy} $\higherpol$.
This policy reuses the set of composable policies ${\compopolSet = \{\compopol\}_{1}^{K}}$ defined in $\tmdpSet$, by using the set $\activationSet = \{\activationvect\}_{1}^{K}$ where $\mathbf{w}^{[k]} \in \Real^\adim$ is an \emph{activation vector} whose components are used to combine each DoF of the action vector.
Therefore, the compound policy is modeled as a two-level hierarchical policy ${\higherpol = \higherpolfcn(\activationSet, \compopolSet)}$.

The generation process of an action from the compound policy involves first obtaining the actions from $\compopolSet$ in the low level of the hierarchy, and then combining them at the high level of the hierarchy.
Instead, we here exploit the assumptions made for the composable policies and propose two formulations of $\higherpolfcn$ for obtaining a policy $\higherpol(\action|\state)$ that is also conditional Gaussian and defined in terms of the means $\mathbf{m}^{[k]}$ and covariances matrices $\mathbf{C}^{[k]}$ of the composable policies.
As a result, an action from the compound policy can be sampled directly from the resulting conditional distribution, $\action \sim \higherpol(\action|\state)$.

The first option for $\higherpolfcn$ is to consider that the action of the compound policy is the convex combination of elements of actions sampled from the composable policies.
As the actions for each composable policy are conditionally independent given the states, and each $\compopol(\action|\state)$ is conditional Gaussian, the resulting action is also normally distributed.
Therefore, the compound policy is 
${\higherpol(\action|\state) = \gauss(\mathbf{m}, \text{diag}(\mathbf{c}))}$, where the components in $\mathbf{m}$ and $\mathbf{c}$ are computed as
\begin{equation}
\label{eq:weighed_linear_combination}
  c_i = \sum_{k=1}^K \left( w_i^{[k]} \ \sigma_i^{[k]} \right)^2 , \quad
  m_i = \sum_{k=1}^K w_i^{[k]} \ m_i^{[k]} ,
\end{equation}
for all $1 \leq i \leq \adim$, with $w_i^{[k]}$, $m_i^{[k]}$, $\sigma_i^{[k]}$ as the corresponding elements of the activation vector, mean vector, and standard deviation vector for the composable policy $\policy^{[k]}$.

The second alternative for modeling $\higherpolfcn$ is to consider that each component $i$ in the resulting action vector is obtained from a product of conditional Gaussians ${\higherpol(a_i|\state) \propto \prod_{k=1}^{K} (\compopol(a_i|\state) )^{w_i^{[k]}}}$.
As a result, the compound policy is also ${\higherpol(\action|\state) = \gauss(\mathbf{m}, \text{diag}(\mathbf{c}))}$ where the components in $\mathbf{m}$ and $\mathbf{c}$ are computed as
\begin{equation}
\label{eq:gaussian_product_combination}
c_i = \left(\sum_{k=1}^K \frac{w_i^{[k]}}{(\sigma_i^{[k]})^2}\right)^{-1} , \quad
m_i = c_i \left(\sum_{k=1}^K \frac{w_i^{[k]}}{(\sigma_i^{[k]})^2}\ m_i^{[k]}\right) ,
\end{equation}
for all $1 \leq i \leq \adim$, with $w_i^{[k]}$, $m_i^{[k]}$, $\sigma_i^{[k]}$ defined as in the first case.

\subsection{Hierarchical Policy and Q-functions Modeling}
\label{sec:model_multiple_policies_and_values}
The composable tasks in $\task$ are formulated as independent RL problems, and thus the corresponding policies are also independent.
In this line, the mean $\mathbf{m}^{[k]}$ and covariance matrix $\mathbf{C}^{[k]}$ \footnote{More specifically a vector of log standard deviations.} that describe a composable Gaussian policy can be obtained from an independent NN.
Nevertheless, we can exploit the assumption that all the tasks in $\tmdpSet$ share the same state space, and therefore use shared layers across all the policies for obtaining features that can be exploited by all the policies.
The parameters of the resulting NN policy are denoted by $\pparams^{\tmdpSet}$, and include both the parameters of each NN policy and the shared layers.

\begin{figure}[!tp]
\vspace*{2ex}
  \centering
  \begin{subfigure}{0.26\textwidth}
    \centering
    \includegraphics[width=1.\linewidth]{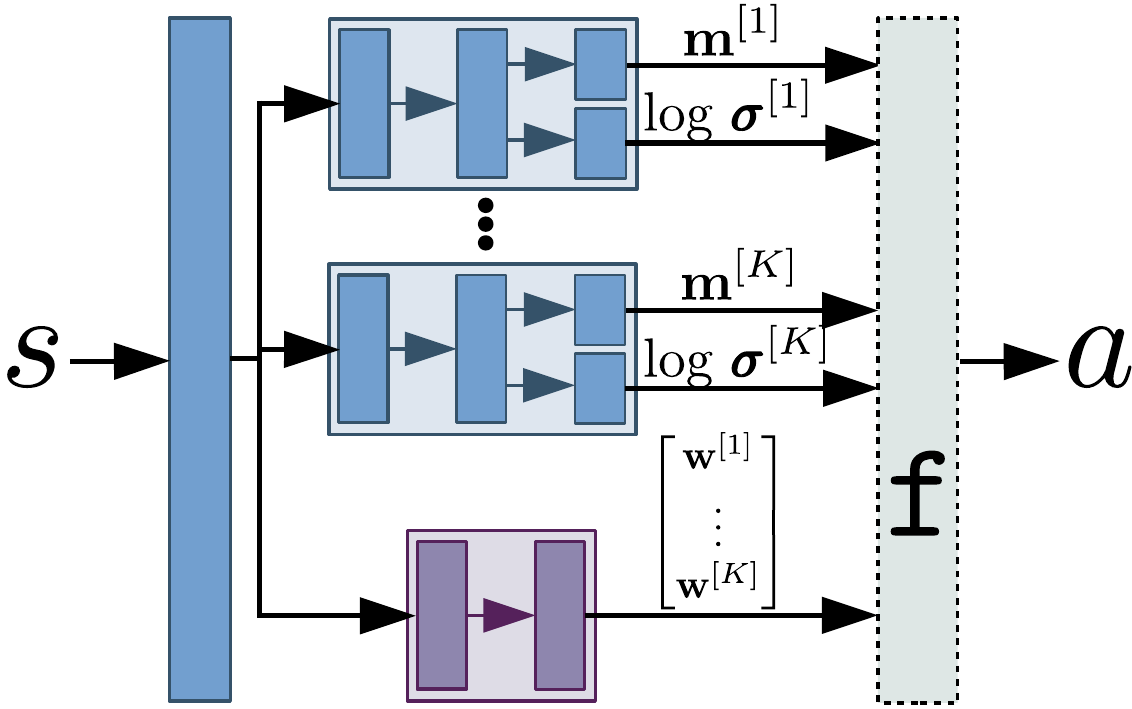}
    \subcaption[]{Hierarchical policy}
    \label{Fig:policy_network}
  \end{subfigure}
  \begin{subfigure}{0.21\textwidth}
    \centering
    \includegraphics[width=.8\linewidth]{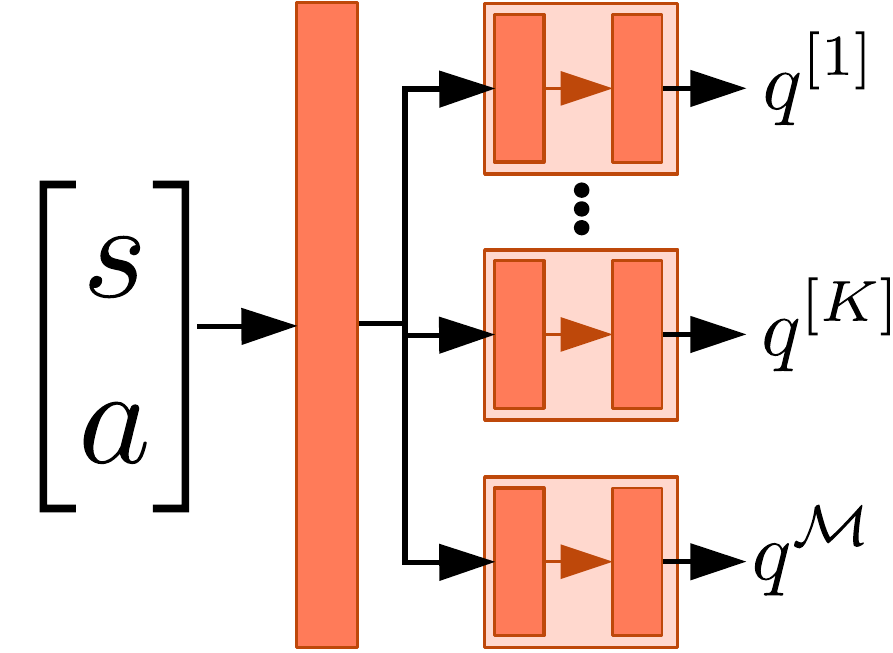}
    \subcaption[]{Q-value function}
    \label{Fig:q_fcn_network}
  \end{subfigure}
  \caption[Policy and Q-value function neural networks]{\textit{Policy and Q-value function neural networks.} \textit{(a)} the first layer is shared among all the modules, the modules for the composable policies are shown in blue and the module that outputs the activation weights is depicted in purple. The outputs of all these modules are then composed in $\higherpolfcn$ by using one of the alternatives described in Section \ref{sec:hierarchical_model}. \textit{(b)} the first layer is also shared among all the modules, and then each module outputs the corresponding approximated Q-value.}  
  \label{Fig:policy_and_q_fcn_networks}
\vspace*{-3ex}
\end{figure}

Moreover, the function $\activationfcn$ required for obtaining the state-dependent activation vectors $\{\mathbf{w}^{[k]}\}_{1}^{K} = \activationfcn(s)$, can also be modeled by an NN with parameters $\pparams^{w}$.
This new NN is included in the previously described NN and thus also exploits the features obtained in the shared layers.
Therefore, the whole hierarchical policy is parameterized by $\ppolicy$, an NN with parameters $\pparams = [\pparams^{\tmdpSet} \quad \pparams^w]$ and depicted in Fig. \ref{Fig:policy_network}.

In the same way, an NN with an architecture similar to the hierarchical policy can be used to model the \hbox{Q-functions} of both the composable policies and compound policy.
Therefore, the resulting NN, denoted by $\pqval$ and depicted in Fig. \ref{Fig:q_fcn_network}, has parameters $\qparams = [\qparams^{\tmdpSet} \quad \qparams^{\mdp}]$.

\section{Simultaneous Learning and Composition of Modular Maximum Entropy Policies}\label{sec:simultanous_learning_and_compo}

Most methods learn the composable tasks one at a time, and later, the compound task.
This procedure is not scalable as all the experience collected for each learning process is only used for that specific process.
Also, it is not possible to start learning more complex tasks unless all the composable policies have been successfully learned.
The method proposed in this section is based on the idea that a single stream of experience can be used to improve not only the policy that is generating the behavior but also, indirectly, many other policies.
Similar to~\cite{cabi17:iu_agent} and~\cite{riedmiller18:sac-x}, our method assumes that the robot receives, at each time step, the rewards for different tasks, and each reward has an assigned policy that tries to maximize its corresponding return $G$ by using the same collection of state-action pairs.


\subsection{Off-Policy Multi-task Policy Search}
The sets of composable policies $\compopolSet$ and activation vectors $\activationSet$ required for a compound policy $\higherpol$ to solve task $\mdp$, can be learned simultaneously.
To do so, let us assume that, at each time step, the robot receives a \emph{stream of rewards} ${\rvect_t = [\rt^{[1]}\;  \dots\;  \rt^{[K]}\; \rt^\mdp]^{\trsp}}$, that is, a vector whose components are the reward $\reward_t^\mdp$ of the compound task $\mdp$ and the reward $\reward_t^{[k]}$ of each composable task in $\tmdpSet$. 

Moreover, the method considers that the \emph{behavior} or \emph{intentional} policy, i.e. the policy followed by the robot to interact with the environment, is always the compound policy $\higherpol$.
The experience at each time step $(\st, \at, \rvect_t, \stp)$ is collected in the dataset $\dataset$, and subsequently used for improving compound and composable policies.
As a result, the data in $\dataset$ is off-policy experience for the composable policies because it is generated from a different policy~\cite{sutton18:RL_an_introduction}.
Thus, the composable policies (from now on \emph{unintentional} policies \cite{cabi17:iu_agent}) are target policies in the off-policy setting.

Therefore, by considering the parameterized policy $\ppolicy$ (see section \ref{sec:model_multiple_policies_and_values}) with parameters ${\pparams = [\pparams^{\tmdpSet} \quad \pparams^w]}$, the optimal parameters $\pparams^{*}$ are those that maximize the objective

\begin{equation}\label{eq:hiu_full_policy_objective}
J_\policy(\pparams) = J_\policy(\pparams^w; \mdp) + \sum_{k=1}^{K}\ J_\policy(\pparams^{\tmdpSet}; \tmdp^{[k]}) ,
\end{equation}
where $J_\policy(\pparams^{\tmdpSet}; \tmdp^{[k]})$ denotes the performance criterion of the composable policy $\policy_{\pparams}^{[k]}$ in task $\task^{[k]}$, and $J_\policy(\pparams^{w}; \mdp)$ the performance criterion of the compound policy $\policy_{\pparams}^{\mdp}$ in task $\mdp$.


\subsection{Multi-task Soft Actor-Critic}
As mentioned in section \ref{sec:max_ent_rl}, the maximum entropy objective incentives exploration, which is critical for the introduced method as the composable policies are learned unintentionally and their influence in the sampling process is indirect.
Thus, each policy seeks to optimize the maximum entropy objective
\begin{equation}\label{eq:maxent_objective_compotasks}
J({\policy^{[j]}}) = \sum_{t=0}^{\infty} \expectation_{\policy^{[j]}} \left[ \discount^t \Big( \reward^{[j]}(\st, \at) + \alpha \entropy(\policy^{[j]}(\cdot | \st)) \Big) \right]
\end{equation}
where $\reward^{[j]}$ is the reward function of the corresponding task $j \in (\tmdp \cup \{\mdp\})$.

Considering the SAC algorithm described in section \ref{sec:sac}, the learning process for all the aforementioned policies is an alternating procedure of policy evaluation, where the value function is computed for all the policies, and policy update, where the policies are improved with their corresponding value functions.
Therefore, at each time step, the parameterized Q-function $\pqval$ optimizes the soft mean-squared bellman error of all the policies
\begin{equation}\label{eq:hiu_qparams_objective}
\resizebox{\hsize}{!}{$J_{Q}(\qparams) = \expectation_{(\st, \at) \sim \rbuffer} \left[\frac{1}{2} \sum_{j}\ \left(Q_{\qparams}^{[j]}(\st, \at) - \left(\reward^{[j]}(\st, \at) + \discount\: \expectation_{\stp \sim \pdyn} [V^{[j]}_{\bar{\qparams}}(\stp)]\right)\right)^2\right]$}
\end{equation}
for all the tasks $j \in (\tmdp \cup \{\mdp\})$,  where the value function $V^{[j]}_{\bar{\qparams}}$ is implicitly parameterized through a target soft Q-function with parameters $\bar{\qparams}$ via
\begin{equation}\label{eq:hiu_value_fcn}
{V^{[j]}_{\bar{\qparams}}(\st) = \expectation_{\at \sim \policy_{\pparams}^{[j]}}[Q^{[j]}_{\bar{\qparams}}(\st, \at) - \alpha \log(\policy_{\pparams}^{[j]}(\at|\st))]}
\end{equation}
On the other hand, the components of \eqref{eq:hiu_full_policy_objective} for the policy improvement step of the parameterized policy $\ppolicy$ are defined as
\begin{equation}\label{eq:hiu_polparams_objective}
\resizebox{\hsize}{!}{$J_\policy(\pparams^\tmdp; \tmdp^{[k]}) = \expectation_{\st \sim \rbuffer}\left[ \expectation_{\action_t \sim \policy_{\pparams}^{[k]}}\left[Q_{\qparams}^{[k]}(\st, \action_t) - \alpha \log(\policy_{\pparams}^{[k]}(\at|\st))\right] \right]$}
\end{equation}
for each composable task $\tmdp^{[k]}$ in $\tmdp$, and
\begin{equation}\label{eq:hiu_compoparams_objective}
\resizebox{\hsize}{!}{$J_\policy(\pparams^w; \mdp) = \expectation_{\st \sim \rbuffer}\left[ \expectation_{\action_t \sim \policy_{\pparams}^{\mdp}}\left[Q_{\qparams}^{\mdp}(\st, \action_t) - \alpha \log(\policy_{\pparams}^{\mdp}(\at|\st))\right] \right]$}
\end{equation}
for the compound task $\mdp$.
Note that the parameters $\pparams^w$ required for modeling the activation vectors in $\activationSet$ are the only ones updated in the compound policy because, as discussed in~\cite{riedmiller18:sac-x}, there is no guarantee to preserve the unintentional policies.
As a consequence, the proposed algorithm improves the parameterized hierarchical policy $\ppolicy$ in a two-step process with random minibatches from $\rbuffer$.
First, by optimizing $\pparams^\tmdp$ with~\eqref{eq:hiu_polparams_objective} for all the composable tasks.
And second, by fixing $\pparams^\tmdp$ and optimizing $\pparams^\mdp$ through \eqref{eq:hiu_compoparams_objective}.

As suggested in~\cite{haarnoja18:sac_algos_and_apps}, the practical algorithm considers two soft Q-function NNs with parameters $\qparams_i$ trained independently to optimize~\eqref{eq:hiu_qparams_objective}.
Furthermore, the algorithm includes a step to calculate $\alpha$ automatically by optimizing 
\begin{equation}
{J(\alpha^{[j]})=\expectation_{\at \sim \policy^{[j]}} \left[- \alpha \log(\policy_{\pparams}^{[j]}(\at|\st)) + \alpha \bar{\entropy}^{[j]}\right]}
\end{equation}
for $j \in (\tmdp \cup \{\mdp\})$.
Therefore, the whole training process with the proposed method, called Hierarchical Intentional-Unintentional (HIU), is summarized in Algorithm \autoref{alg:HIU-SAC}. 

{
\alglanguage{pseudocode}
\begin{algorithm}[t]
  \footnotesize
  \caption{\strut HIU-SAC}\label{alg:HIU-SAC}
  \begin{algorithmic}[1]
    \State Initialize target network weights: $\bar{\qparams}_i \leftarrow \qparams_i$ for $i \in \{ 1, 2\}$
    \State Initialize an empty replay memory $\rbuffer \leftarrow \emptyset$
    \For{each iteration}
    
    \For{each interaction step}
    \State Sample compound action $\at \sim \ppolicy(\at|\st)$
    \State  \begin{varwidth}[t]{\linewidth}
      Sample transition from the  environment: ${\stp \sim \pdyn(\stp | \st, \at)}$
      \end{varwidth}
    \State \begin{varwidth}[t]{\linewidth}
      Store the interaction data in the replay memory: \par
      ${\rbuffer \leftarrow \rbuffer \cup \{ (\st, \at, \rvectt, \stp)\}}$
      \end{varwidth}
    \EndFor
    
    \For{each gradient step}
    \State \begin{varwidth}[t]{\linewidth}
      Update $Q$-networks parameters:$\qparams_i \leftarrow \lambda_{Q} \hat{\nabla} J_{Q}(\qparams_i)$ for $i \in \{ 1, 2\}$
    \end{varwidth}

    \State \begin{varwidth}[t]{\linewidth}
      Update composable policies parameters: ${\pparams^\tmdp \leftarrow \lambda_{\policy} \hat{\nabla} J_\policy(\pparams^\tmdp; \tmdpSet)}$
      \end{varwidth}
    \State \begin{varwidth}[t]{\linewidth}
      Update compound policy parameters: ${\pparams^\mdp \leftarrow \lambda_{\policy} \hat{\nabla} J_\policy(\pparams^\mdp; \mdp)}$
      \end{varwidth}

    \State \begin{varwidth}[t]{\linewidth}
      Update temperature parameters: \par
      $\alpha^{[j]} \leftarrow \lambda_{\alpha} \hat{\nabla} J_{\alpha}(\alpha^{[j]})$ for $j \in (\tmdp \cup \{\mdp\})$
    \end{varwidth}
    \State \begin{varwidth}[t]{\linewidth}
      Update target Q-networks weights: \par
      ${\bar{\qparams_i} \leftarrow \rho \qparams_i + (1 - \rho) \bar{\qparams_i}}$ for $i \in \{ 1, 2\}$
    \end{varwidth}
    \EndFor   
    \EndFor
  \end{algorithmic}
\end{algorithm}
}

\section{Experiments}\label{sec:Experiments}
In order to analyze the proposed approach, several experiments were carried out in four robotic tasks that can be intuitively decomposed into simpler tasks.
The goal of these experiments is to evaluate if the proposed approach \textit{1)} solves an RL problem with a policy that reuses a set of composable policies, and at the same time, \textit{2)} obtains composable policies with performance similar to dedicated single-task policies.

\subsection{Tasks Description}\label{sec:TasksDescription}
\begin{figure}[!tp]
  \vspace*{1ex}
  \centering
  \begin{subfigure}{0.21\textwidth}
    \centering
    \includegraphics[width=0.9\linewidth]{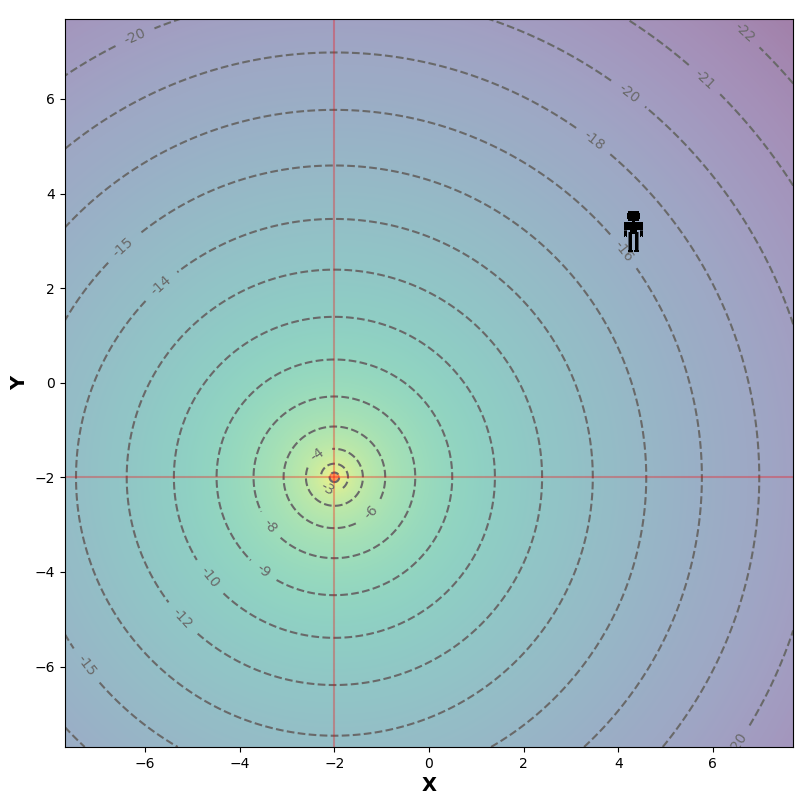}
    \vspace*{-1.0ex}
    \subcaption[]{}
    \label{Fig:description_navigation2d}
  \end{subfigure}%
  \begin{subfigure}{0.21\textwidth}
    \centering
    \includegraphics[width=0.6\linewidth]{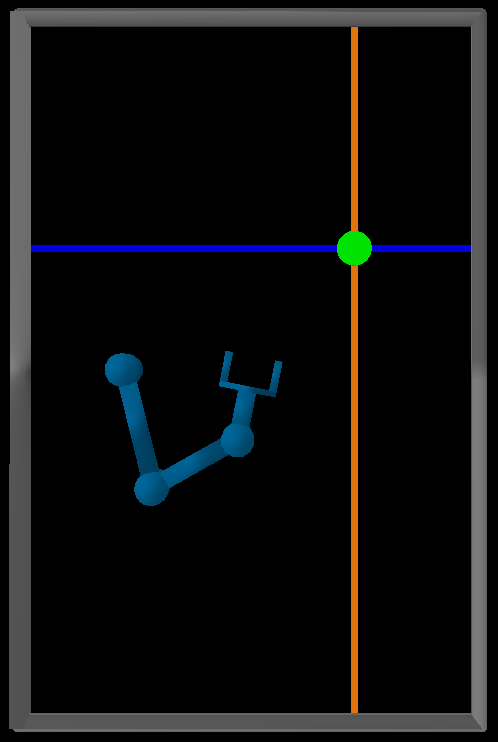}
    \vspace*{-1.0ex}
    \subcaption[]{}
    \label{Fig:description_reacher}
  \end{subfigure}

  \vspace*{1.0ex}

  \begin{subfigure}{0.21\textwidth}
    \centering
    \includegraphics[width=0.6\linewidth]{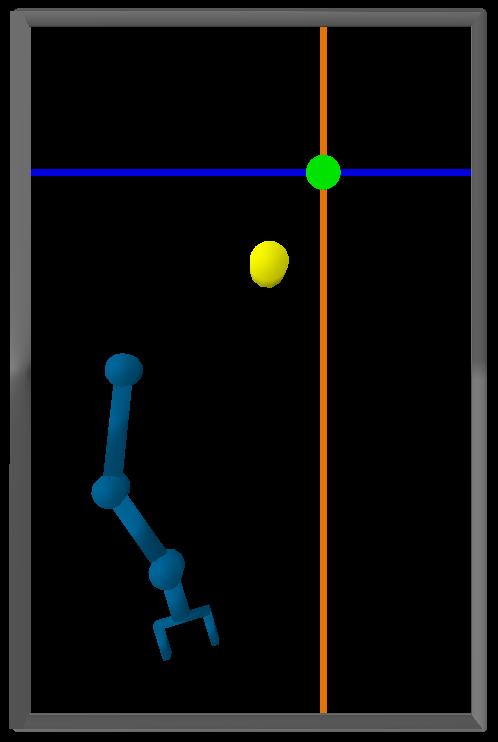}
    \vspace*{-1.0ex}
    \subcaption[]{}
    \label{Fig:description_pusher}
  \end{subfigure}%
  \begin{subfigure}{0.21\textwidth}
    \centering
    \includegraphics[width=0.9\linewidth]{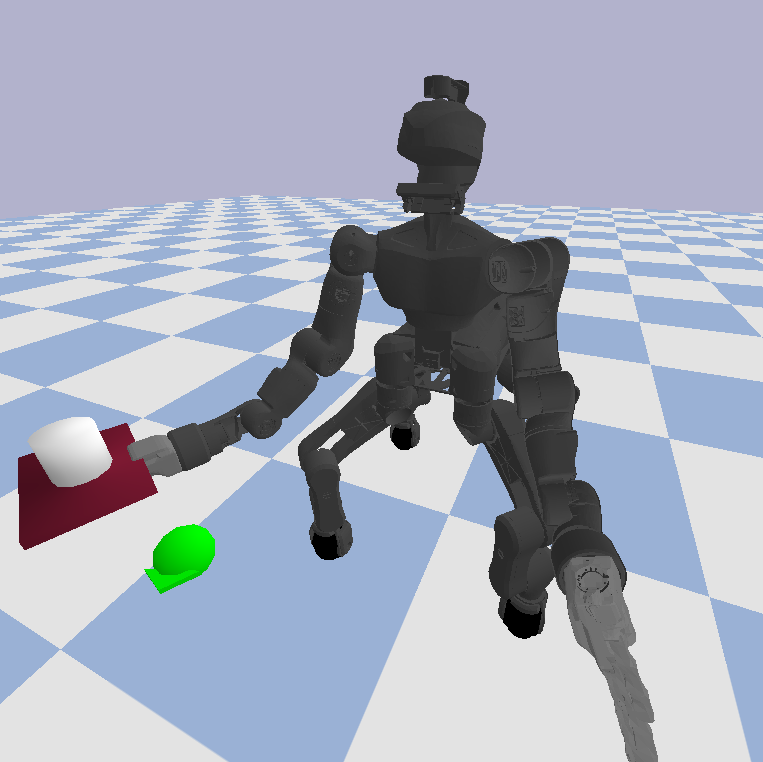}
    \vspace*{-1.0ex}
    \subcaption[]{}
    \label{Fig:description_centauro}
  \end{subfigure}
  \caption[Environments considered in the experiments]{\textit{Experimental scenarios:}
    (\subref{Fig:description_navigation2d}) 2D particle that reaches a fixed goal position.
    (\subref{Fig:description_reacher}) 3-DoF planar manipulator reaching a random 2D goal position (green circle).
    (\subref{Fig:description_pusher}) 3-DoF planar manipulator that should push the randomly-placed yellow cylinder to a varying 2D goal position (green circle).
    (\subref{Fig:description_centauro}) CENTAURO robot that simultaneously balances a tray with a cylinder and moves to a random 3D goal pose.
    } 
  \label{Fig:description_envs}
  \vspace*{-1ex}
\end{figure}

In the first environment, shown in \mbox{Fig. \ref{Fig:description_navigation2d}}, the agent is a 2D point particle that has to reach the position $(-2, -2)$.
The state of this environment is continuous and defined by the position $(x, y)$ of the particle, and the control actions are its velocities $(\dot{x},\dot{y})$, then $\sdim = \adim = 2$.
The initial position of the particle is sampled from a spherical Gaussian distribution centered in the position $(4, 4)$.
This task can be decomposed into two composable tasks, namely, reaching the position $-2$ in the $x$ coordinate, and reaching the position $-2$ in the $y$ coordinate.
Therefore, the compound policy to reach $(-2, -2)$ has to combine the corresponding composable policies.

The second and third environments correspond a 3-DoF planar manipulator simulated in Pybullet~\cite{coumans18:pybullet}, and whose control actions are joint torques, then $\adim=3$.
The second environment, shown in \autoref{Fig:description_reacher}, requires the robot to reach a random goal pose and is described by a state composed of the joint positions and velocities of the robot, and the relative position of the robot end-effector w.r.t the goal, then, $\sdim = 8$.
The third environment, displayed in \autoref{Fig:description_pusher}, requires the manipulator to reach a cylinder and push it to a target location.
In addition to the joint positions and velocities, the state also includes the positions of the end-effector, the cylinder and the goal, then $\sdim = 12$.

Finally, the proposed approach is also tested in a more complex task, where a simulated CENTAURO robot~\cite{baccelliere17:centauro} has to reach a target 3D pose while balancing an object with a tray, as depicted in \autoref{Fig:description_centauro}.
The tray is firmly attached to the robot hand but not the cylinder.
The control actions are task joint torques of the right arm, then $\adim = 7$.
Note that if a zero-torque control action was applied to the joints, the cylinder would fall by its weight.
The state in this scenario is composed of the arm joint positions and velocities, pose errors and rate of change of the errors between the target pose and the center of the tray, and between the desired pose of the cylinder in the tray and its current pose, then $\sdim = 38$.

\subsection{Robot Learning Details}\label{sec:TrainingDetails}

The NN models proposed in \autoref{sec:model_multiple_policies_and_values} are used for learning the four tasks above described.
The same architecture is used in all the experiments, namely, NNs with ReLU nonlinearities in the hidden nodes and none in the outputs.
However, the number of nodes depends on the task, as summarized in~\autoref{table:experiments_details}.

The tasks are learned with the algorithm proposed in \autoref{sec:simultanous_learning_and_compo} and the following hyperparameters: Adam optimizer with learning rates $\lambda_{Q} = \lambda_{\policy} = \lambda_{\alpha} = 3\cdot10^{-4}$, target smoothing coefficient $\rho = 5\cdot10^{-3}$, and discount factor $\discount = 0.99$.
The NNs are trained using stochastic gradient descent with batches sampled from $\rbuffer$ after an interaction with the environment.
The entropy target $\bar{\entropy}^{[j]}$ is the same for both composable and compound policies, however its value varies for each task.
These values and the replay buffer size $\rbuffer$ for each environment are also depicted in~\autoref{table:experiments_details}.

The two composition strategies described in \autoref{sec:composition_max_ent_policies} are denoted with HIUSAC.
The alternative that considers \autoref{eq:weighed_linear_combination} is denoted by HIUSAC-1,  while the solution using \autoref{eq:gaussian_product_combination} is denoted by HIUSAC-2.
For comparison purposes, the SAC algorithm was used to learn both compound and composable policies in single-task RL formulations. 

\begin{table}[!htb]
  \renewcommand{\arraystretch}{1.3}
  \caption{Environment-specific hyperparameters}
  \label{table:experiments_details}
  \centering
  \begin{tabular}{ccccc}
    & (1) & (2) & (3) & (4)\\
  \hline
  
  Units per layer & 64 & 128 & 128 & 256 \\
  Training steps & $1.5\! \cdot\! 10^{4}$ & $1.5\! \cdot\! 10^{5}$ & $1.5\! \cdot\! 10^{5}$ & $1.5\! \cdot\! 10^{6}$ \\
  Size $\rbuffer$ & $5\! \cdot\! 10^{6}$ & $5\! \cdot\! 10^{6}$ & $5\! \cdot\! 10^{6}$ & $1\! \cdot\! 10^{7}$ \\
  Size Minibatch & 64 & 256 & 256 & 256 \\
  $\bar{\entropy}$ & 0 & 1 & 1 & 1 \\
  \hline
  \end{tabular}
  \vfill
  \vspace{2mm}
  \raggedright
  {
    (1) 2D navigation with point particle (\autoref{Fig:description_navigation2d}). \\
    (2) Reaching task with 3 DoF manipulator (\autoref{Fig:description_reacher}) \\
    (3) Pushing task with 3 DoF manipulator (\autoref{Fig:description_pusher}) \\
    (4) Reaching and tray balancing task with CENTAURO (\autoref{Fig:description_centauro})
  }
\end{table}

\subsection{Results}\label{sec:Results}

\begin{figure}[!htb]
  \vspace*{2ex}
  \centering
  \begin{subfigure}[b]{.48\textwidth}
    \centering
    \includegraphics[width=\linewidth]{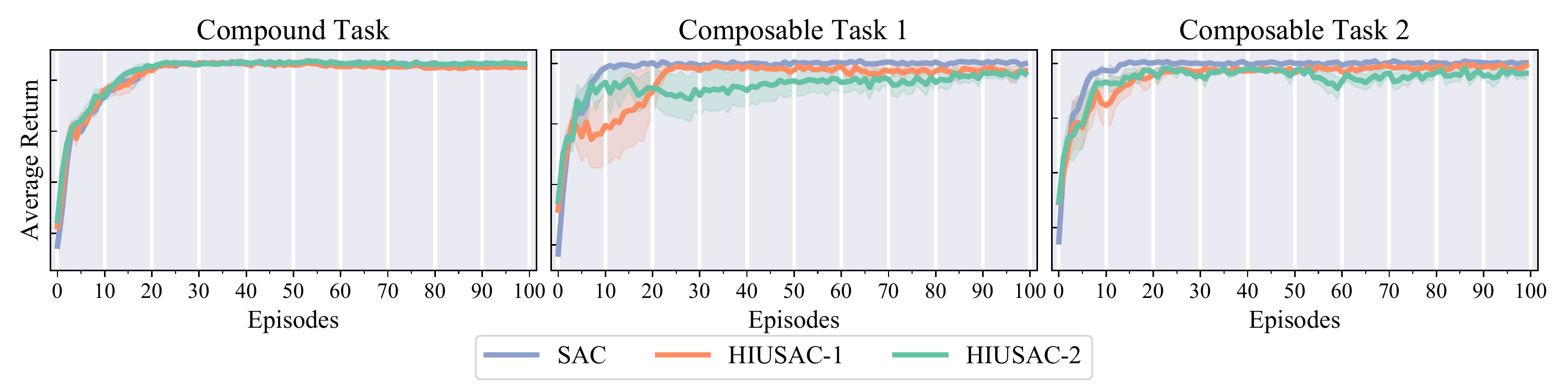}
    \vspace*{-4.0ex}
    \subcaption[]{}
    \label{Fig:learning_navigation2d}
  \end{subfigure}
    
  \vspace*{1.0ex}

  \begin{subfigure}[b]{.46\textwidth}
    \centering
    \includegraphics[width=\linewidth]{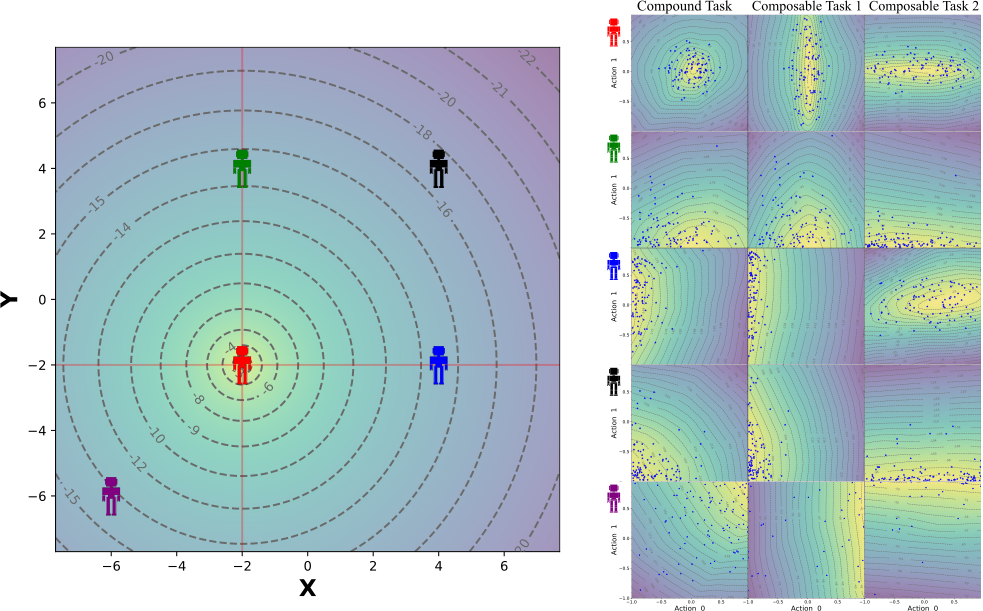}
    \subcaption[]{}
    \label{Fig:values_navigation2d}
  \end{subfigure}
  \caption{\textit{Navigation task of a 2D point particle:} 
  (a) SAC denotes the compound and composable policies obtained with the SAC algorithm in single-task formulations. The policies obtained with the algorithm proposed in \autoref{sec:composition_max_ent_policies} are denoted by HIUSAC-1 and HIUSAC-2, with the former considering \autoref{eq:weighed_linear_combination} and the latter using \autoref{eq:gaussian_product_combination}. The learning curves show that both the compound and composable policies can successfully perform their respective tasks using the proposed hierarchical model, while being competitive with the single-task formulations. (b) Actions sampled from these policies seek to move the point particle to their respective target. The soft Q-values of these policies, depicted as contour plots, show the values for the actions in five different positions.}
  \vspace*{2ex}
\end{figure}

\autoref{Fig:learning_navigation2d} shows the learning curves of the composable policies obtained with both HIUSAC-1 and HIUSAC-2 for the 2D particle environment.
The achieved performance is similar to that obtained directly in the compound task with the SAC algorithm.
The approximated soft Q-values for the velocities (actions) given some specific positions of the particle (state) are depicted in \autoref{Fig:values_navigation2d}.
Notice how the actions and soft Q-values vary as a function of the position, capturing the specifications of their respective tasks.
This is a remarkable result as the composable policies were learned unintentionally with off-policy experience collected only with the compound policy.

\begin{figure*}[!htp]
  \vspace*{1ex}
  \centering
  \begin{subfigure}{.72\textwidth}
    \centering
    \begin{subfigure}{\textwidth}
      \centering
      \includegraphics[width=\linewidth]{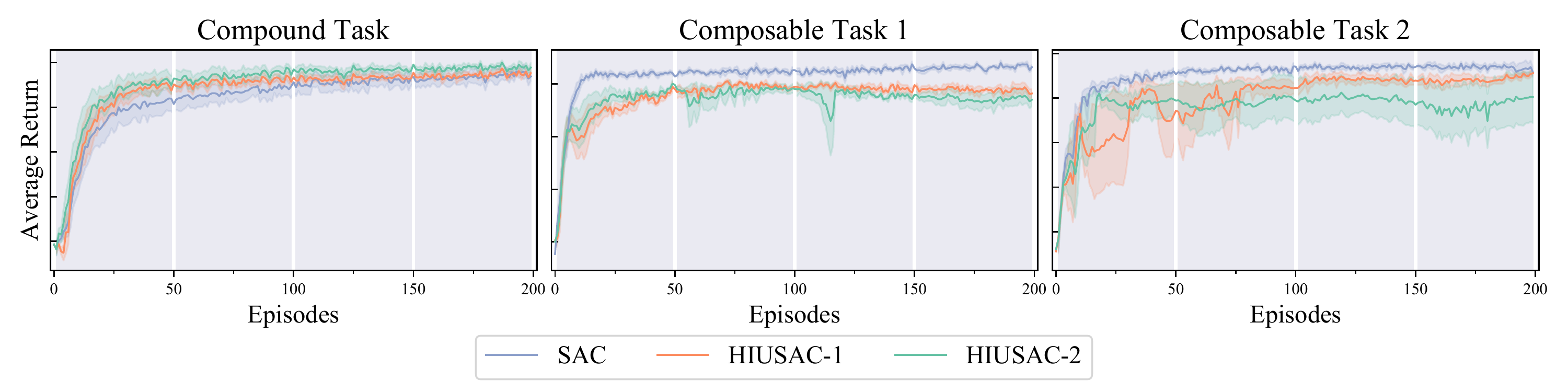}
    \end{subfigure}
    \vspace*{-1.5ex}
    \subcaption[]{Reaching Environment}
    \label{Fig:learning_reaching}
  \end{subfigure}

  \vspace*{1.0ex}
  
  \begin{subfigure}{.72\textwidth}
    \centering
    \begin{subfigure}{\textwidth}
      \centering
      \includegraphics[width=\linewidth]{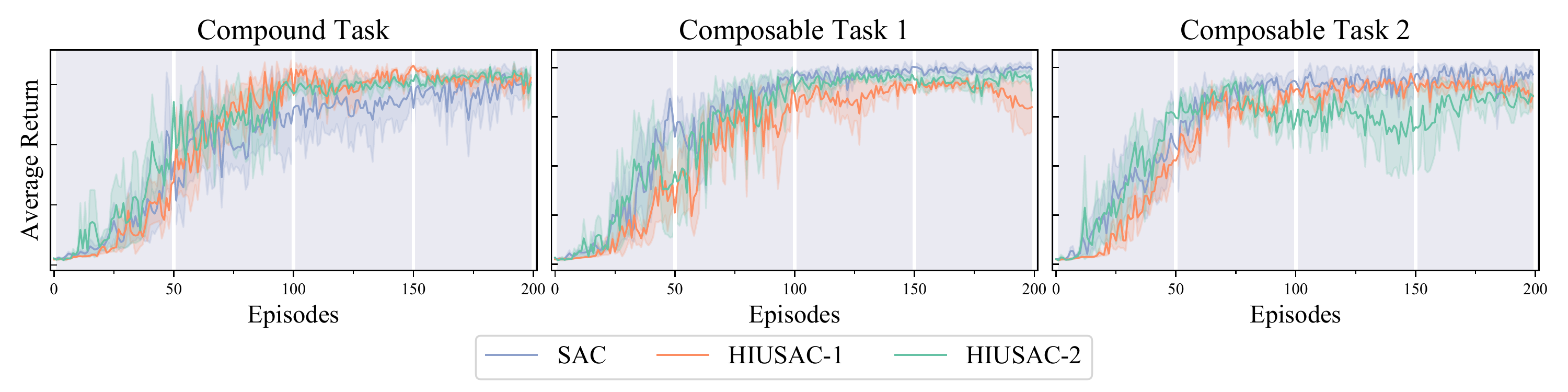}
    \end{subfigure}
    \vspace*{-1.5ex}
    \subcaption[]{Pushing Environment}
    \label{Fig:learning_pushing}
  \end{subfigure}
  
  \vspace*{1.0ex}
  
  \begin{subfigure}{.72\textwidth}
    \centering
    \begin{subfigure}{\textwidth}
      \centering
      \includegraphics[width=\linewidth]{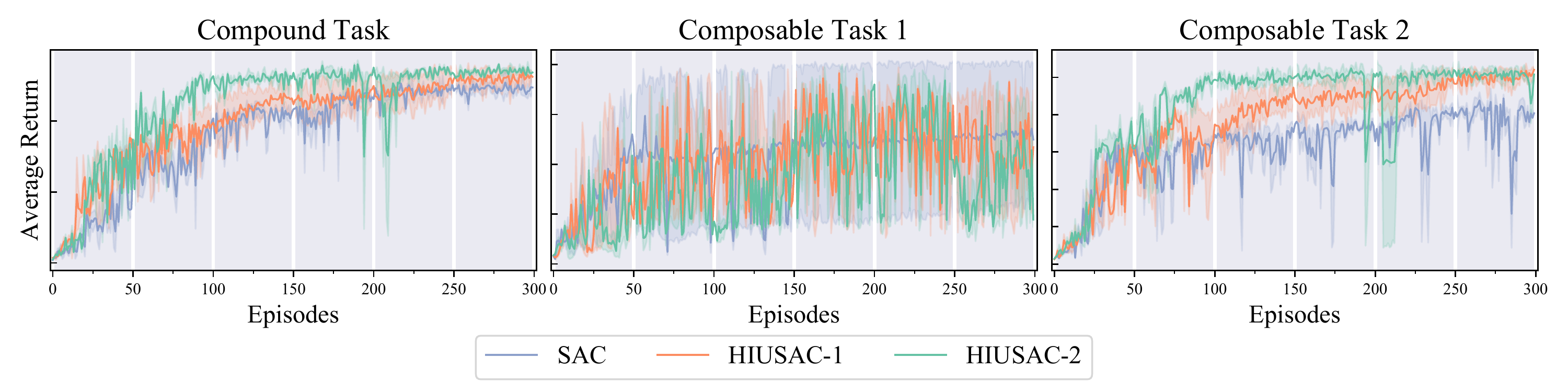}
    \end{subfigure}
    \vspace*{-1.5ex}
    \subcaption[]{CENTAURO Environment}
    \label{Fig:learning_centauro}
  \end{subfigure}
  
  \caption{\textit{Learning curves for the compound task and the composable tasks for the reaching, pushing and CENTAURO tasks.} The figures show the average return for the proposed approach with the two composition strategies described in \autoref{sec:composition_max_ent_policies}. The method that considers \autoref{eq:weighed_linear_combination} is denoted by HIUSAC-1. On the other hand, the method that considers \autoref{eq:gaussian_product_combination} is denoted by HIUSAC-2 The resulting policies obtained with both alternatives are compared with the ones obtained from single-task RL formulations with the SAC algorithm.}
  \label{Fig:learning_all_envs}
\end{figure*}

The learning curves of the other three environments are displayed in \autoref{Fig:learning_all_envs}.
As noted previously, the tasks with the planar manipulator are more complex than the navigation of the 2D particle because the action space, i.e. joint torques, influence directly in the task of reaching the $x$ position and the task of reaching the $y$ position.
Therefore, it is more difficult to assign proper activation vectors for the respective composable policies.
However, both HIUSAC-1 and HIUSAC-2 obtained successful composable and compound policies, all of them with a performance similar to the policies obtained  with the SAC algorithm in single-task RL formulations.
However, as we can notice in the composable task 2 for the reaching environment (\autoref{Fig:learning_reaching}), both alternatives require more iterations to converge.

Finally, the results obtained for the task carried out by the CENTAURO robot are reported in \autoref{Fig:learning_centauro}.
In this case, the composable policy converges faster and results in higher average returns when compared to the policy obtained in the single-task formulation.
This results demonstrate how the complexity of one task can be solved with a collection of simpler subtasks by exploiting a hierarchical off-policy formulation.
An interesting result is that the performance of the composable policies for task 2 exceeds that of their single-task counterparts.
We attribute this to the fact that the compound policy explores better the environment and therefore the collected experience contains more meaningful information than those obtained in single-task RL formulations.
Between HIUSAC-1 and HIUSAC-2, the latter converges faster and results in higher average returns in the compound task and also the composable ones.

\section{Conclusions and Future Work}\label{sec:Conclusions}
In this paper we have proposed a hierarchical RL approach for tasks that can be decomposed into a collection of subtasks that require to be performed concurrently.
The Gaussian policies corresponding to these subtasks are combined using a set of activation vectors.
These activation vectors allow to consider concurrently actions sampled from all the low-level policies and preferences among specific components.
Furthermore, two methods were proposed to obtain a compound policy that is also Gaussian and a function of the means and covariances matrices of the composable policies. 

Moreover, we proposed an algorithm for learning both compound and composable policies within the same learning process by exploiting the off-policy data generated from the compound policy.
Note that populating the replay memory buffer with rich experiences is essential for acquiring multiple skills in an off-policy manner.
The composable policies learned unintentionally had similar performance than the policies obtained in single-task formulations only when the compound policy was able to efficiently explore the environment.
For this reason, the algorithm was built on a maximum entropy RL framework to favor exploration during the learning process.

Nevertheless, choosing the temperature parameters for the maximum entropy RL objective is challenging for the compound policy because its stochasticity is determined by the activation vectors and the stochasticity of all the composable policies.
As a result, high temperature values favors higher entropy policies and the compound policy will show preference for composable policies with high stochasticity.
However, this preference is made to the detriment of preferring policies with good performance but low entropy.
We have used the automating entropy adjustment strategy proposed in \cite{haarnoja18:sac_algos_and_apps} that reduces this problem, however now the problem is derived to choose the minimum expected entropy for both composable and compound policies.
This value was easier to set for the simpler environments of the experiments, but more challenging for the complex ones.
Therefore, future work could be focus in developing a mechanism that allows to obtain this value automatically based on the performance of both compound and composable policies.

On the other hand, in this paper, the action value functions of the composable policies are only used in the policy evaluation step of these policies.
However, these models capture the performance of the policies in their respective tasks, and therefore they are important sources of information that could be considered by the compound policy. 

Finally, an important motivation for task decomposition is reusing the composable policies in new tasks.
Future work will be to analyze the behavior of the composable policies when they are reused in different compound tasks.
It is important to know how the performance of new compound policies is affected by composable policies obtained in different contexts.
Moreover, the experiments carried out in this paper were focused in tasks that are decomposed in two subtasks.
Future work will consider tasks that could be decomposed in more than two subtasks and tasks where the components of the action vector can be assigned completely to specific tasks, e.g. bimanual tasks.




%

\section*{ACKNOWLEDGMENT}
The TITAN Xp used for this research was donated by the NVIDIA Corporation.

\bibliographystyle{IEEEtran}
\bibliography{hiu.bbl} 

\end{document}